\pdfoutput=1
\documentclass[]{trbunofficial}
\usepackage{graphicx}

\usepackage{booktabs}

\usepackage[hidelinks]{hyperref}

\AuthorHeaders{Cai, and Wang}
\title{From Trial to Deployment: A SEM Analysis of Traveler Adoptions to Fully Operational Autonomous Taxis}

\author{%
  \textbf{Yutong Cai, Ph.D.}\\
  Engineering Systems and Design, Singapore University of Technology and Design\\
  yutong\_cai@sutd.edu.sg\\
  \hfill\break
  \textbf{Hua Wang, Ph.D., Corresponding Author}\\
  School of Automotive and Transportation Engineering, Hefei University of Technology\\
  hwang191901@gmail.com
}

 \TotalWords{5033}

\begin{document}
\maketitle

\section{Abstract}

Autonomous taxi services represent a transformative advancement in urban mobility, offering safety, efficiency, and round-the-clock operations. While existing literature has explored user acceptance of autonomous taxis through stated preference experiments and hypothetical scenarios, few studies have investigated actual user behavior based on operational AV services. This study addresses that gap by leveraging survey data from Wuhan, China, where Baidu's Apollo Robotaxi service operates at scale. We design a realistic survey incorporating actual service attributes and collect 336 valid responses from actual users. Using Structural Equation Modeling, we identify six latent psychological constructs, namely Trust \& Policy Support, Cost Sensitivity, Performance, Behavioral Intention, Lifestyle, and Education. Their influences on adoption behavior, measured by the selection frequency of autonomous taxis in ten scenarios, are examined and interpreted. Results show that Cost Sensitivity and Behavioral Intention are the strongest positive predictors of adoption, while other latent constructs play more nuanced roles. The model demonstrates strong goodness-of-fit across multiple indices. Our findings offer empirical evidence to support policymaking, fare design, and public outreach strategies for scaling autonomous taxis deployments in real-world urban settings.

\hfill\break%
\noindent\textit{Keywords}: Autonomous vehicle, Autonomous taxi, SP survey, Structural equation modelling
\newpage

\section{Introduction}
In recent years, autonomous vehicles (AVs) have emerged as a transformative technology with the potential to redefine personal mobility. Their appeal lies in their capacity to improve road safety, enhance traffic efficiency, and reduce the need for human drivers. This has also enabled new possibilities for shared mobility and on-demand services. Among various AV applications, autonomous taxi (AT) services have garnered increasing attention from both industry stakeholders and transportation planners due to their potential to offer affordable, labor-free, and round-the-clock urban mobility. An estimation by \citep{jiang2023diffusion} revealed that the AT services will likely to dominate the mobility market by taking 98\% mode share before the year 2057.

Despite this growing interest, most empirical studies assessing public acceptance and behavioral responses toward AT services have relied on hypothetical or simulated scenarios (e.g., \cite{vosooghi2019robo, lee2022effect, yao2025transition}). These include stated preference experiments, vignette-based surveys, and even immersive virtual reality setups, often involving respondents who have never used an autonomous taxi before to state their travel mode choices based on their imaginations. Such approaches may fail to capture actual user perceptions and adoption behavior, thus introducing cognitive biases between imagined and experienced services. This gap in external validity presents a challenge for accurately understanding the real-world behavioral drivers of AT adoption.

This study addresses that gap by investigating the adoption behavior of ATs based on the actual deployment of Apollo Robotaxi (AR), which is an autonomous mobility-on-demand (MoD) service developed by the Chinese company Baidu classified as SAE Level 4/5 \href{https://www.sae.org/blog/sae-j3016-update}. Since August 2024, AR has operated fully driverless services in multiple Chinese cities, completing hundreds of thousands of trips with a fleet approaching 1,000 vehicles \citep{EvelynCNBC}. Its large-scale, public-facing operations offer an unprecedented opportunity to study AT adoption grounded in real experience rather than imagined scenarios.

Our study focuses on Wuhan, the first city to maintain continuous, full-scale AR operations with substantial ridership. The AR service disrupted the local mobility market by offering highly subsidized fares, which is roughly one-third of traditional taxi rates. Such aggressive pricing has generated widespread public engagement as well as controversy over its economic and regulatory impacts. This unique context allows us to examine the perceptions and behavioral responses of travelers who have interacted with fully operational AR services firsthand.

Using data collected from both AR users and general travelers in Wuhan, we apply Structural Equation Modelling (SEM) to identify and measure the latent psychological constructs that influence adoption behavior. In particular, we aim to answer the research question of "What are the key latent factors that shape travelers’ behavioral intentions to adopt autonomous taxis?"

Through this approach, we move beyond descriptive statistics or revealed preferences to uncover the underlying attitudinal dimensions that drive AT adoption. The SEM framework allows us to test not only the direct effects of these constructs on behavioral intention but also their interrelationships, thereby offering a holistic view of public acceptance dynamics in the context of operational autonomous mobility services.

This research provides valuable empirical evidence for policymakers and mobility operators seeking to promote the sustainable deployment of ATs. It contributes to a more grounded understanding of public attitudes by leveraging real-world service exposure and rigorous latent construct modeling. Ultimately, our findings inform regulatory design and marketing strategies surrounding AV integration in urban transportation systems from the perspective of policy makers and AT companies, respectively.

The rest of the paper is organized as follows. Section 2 provides a comprehensive review of relevant literature. In Section 3, we present the research framework and the tailored rank-ordered logit models. Section 4 elaborates the questionnaire design and data collection for our SP survey. In Section 5, we present the model estimation results and their policy implications. The last section concludes the paper.

\section{Literature Review}
\label{sec: liter}
The AT services have long been considered as a game changer for the urban transportation systems, with all the advantages from the autonomous nature of the vehicles such as travel cost reduction (less reliant on manpower and higher vehicle utilization rates), traffic congestion alleviations, and road spaces savings \citep{narayanan2020shared}. In the existing literature, a substantial body of work has investigated user attitudes toward AVs, particularly in the context of privately owned AVs (e.g., \citep{bansal2016AVAustin, faisal2023understanding}) as well as their integration into ride-hailing platforms (e.g., \citep{Contreras:2018, Chen:2024, Gao:2024}). These studies have primarily focused on assessing public acceptance and identifying key determinants of AV adoption such as perceived usefulness, trust, safety, socio-demographic characteristics, and willingness to pay (WTP). The insights gained from these investigations are valuable for informing service design and forecasting behavioral responses in future mobility systems \citep{cai2019investigating}.

Moreover, a recent review by \citep{zhang2024evolution} underscores a shift in scholarly attention toward shared AV services, reflecting growing interest in mobility-on-demand (MoD) applications. In this context, our literature review specifically concentrates on research exploring the adoption and behavioral dynamics associated with AV-based MoD services (i.e., ATs), which are central to understanding the evolving role of autonomous mobility in urban transport ecosystems.

With the growing presence of ATs and their expected integration into future transport systems \citep{masoud2017autonomous, yang2023trip}, a substantial body of research has emerged aiming to understand travelers’ mode choice behavior for AT services in the new travel mode basket. These studies, grounded primarily in discrete choice modelling (DCM) frameworks, focus on quantifying how alternative-specific attributes (e.g., travel cost, waiting time, travel time) and individual socio-demographic characteristics (e.g., income, age, education level) influence the likelihood of choosing AT services \citep{jiao2022incentivizing}. Much like earlier work on ride-hailing adoption, this line of inquiry emphasizes observable utility factors that drive mode preferences.

However, while such models provide valuable insights into what variables affect mode choice decisions, they often overlook the underlying cognitive and attitudinal constructs that shape these preferences. In particular, there is limited understanding of the psychological foundations behind the estimated utility coefficients including trust in automation, perceived safety, or policy acceptance that are likely to play a critical role in the adoption of a novel mode like ATs. Bridging this gap requires moving beyond conventional DCM approaches to incorporate latent constructs that represent travelers' beliefs, perceptions, and behavioral intentions. Investigating these latent constructs can offer a deeper explanation of why certain mode attributes matter more to different individuals and reveal the inherent cognitive drivers of AT adoptions. Therefore, apart from the DCMs on calibrating the travel utility functions in the mode choice involving the new AT mode, another research direction on investigating the adoption intention of the AT services uses structural equation modelling (SEM) to unveil the relationships among potential variables that affect travelers’ willingness to switch from existing travel modes to use the new AT services. 

\citep{hassan2019factors} used SEM to investigate the complex relationships between the variables and identify the latent constructs of AV adoption for older Canadians. Based on a national survey, they revealed that older adults’ willingness to adopt AVs was positively associated with factors such as frequent use of alternative transport modes (e.g., public transit, ride sharing), greater travel distance, higher income, being male, and living in urban areas. Conversely, strong dependence on personal driving is negatively associated with AV acceptance. Compared to younger adults, older Canadians express more skepticism about AV technology, with many preferring personal control over their vehicles and showing reluctance to pay extra for AV features. The findings underscore the importance of considering demographic and behavioral characteristics when planning for AV integration, particularly to address the mobility needs of an aging population. \citep{lee2022effect} examined how real-world user experience with ATs affects user acceptance by analyzing data from a field test conducted in South Korea. The study also employs SEM to explore the relationships among user experience, perceived usefulness, trust, and behavioral intention. Findings show that a positive user experience significantly enhances trust and perceived usefulness, which in turn drive the intention to use AT services. Notably, direct experience with the service reduces psychological barriers and increases openness to future adoption. The study highlights the critical role of firsthand experience in shaping public perception and acceptance of autonomous mobility services. \citep{yao2025transition} examined the factors influencing passenger intentions to transition from traditional ride-hailing services to the AT services using the theory of planned behavior. Based on survey data collected from first-tier Chinese cities, the study identifies social anxiety and perceived safety as critical factors significantly influencing users' transition intentions. Specifically, social environment anxiety emerged as the most impactful dimension of social anxiety, while travel habits moderated the relationship between subjective norms, perceived behavioral control, and transition intentions. The findings offer valuable insights for urban planners and AT service providers, emphasizing the need to consider psychological factors and habitual travel behaviors when introducing autonomous mobility services in urban areas.

Collectively, the aforementioned studies underscore the importance of integrating advanced modeling techniques with survey-based data to analyze user preferences for AT services. These investigations highlight key determinants such as cost, travel time, user perceptions, trust, perceived safety, and socio-demographic attributes, forming a critical foundation for designing effective strategies to enhance adoption and optimize AT services. However, a common limitation among these studies is their reliance on hypothetical AT attributes and simulated operational scenarios, often requiring respondents to select travel modes based on speculative or imagined conditions. 

Nevertheless, these methodological improvements still do not fully replicate the complexity and nuances of real-world user behavior, particularly the underlying latent psychological constructs influencing AT adoption in contexts where AV MoD services are actively and widely operational. This study addresses this gap by employing SEM to analyze traveler preferences and behavioral intentions within the context of the established Apollo Robotaxi (AR) service in Wuhan, China. By examining genuine user experiences, real service attributes, and interactions with actual operational environments, this research provides deeper insights into latent constructs such as perceived reliability, social influence, trust, performance expectations, and policy support. The investigation specifically explores how these psychological dimensions shape adoption behaviors, going beyond observable variables to uncover the cognitive foundations of mode choice decisions. The findings substantially contribute to a more nuanced and informed transition toward widespread urban AV mobility.

\section{Methodology}
\label{sec: method}

\begin{figure}[!ht]
  \centering
  \includegraphics[width=0.6\textwidth]{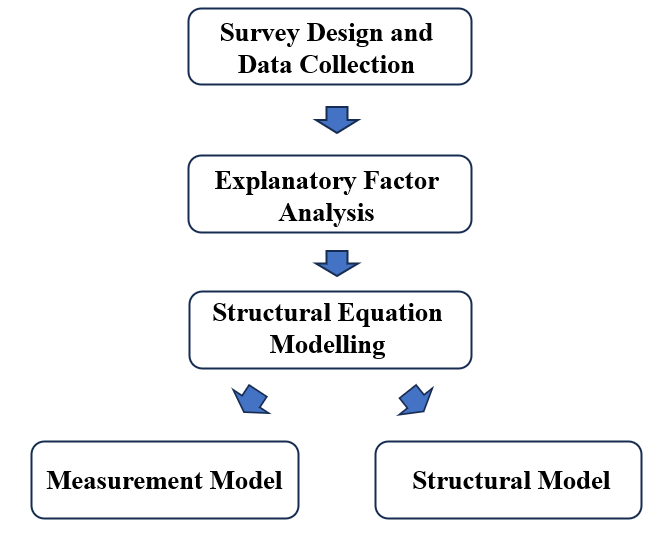}
  \caption{An overview of our research framework}\label{fig: framework}
\end{figure}

Our research framework is illustrated in Figure \ref{fig: framework}. To deeply understand traveler perceptions toward the reshaped mobility market and adoption behaviors of the new AT service AR, we first design a SP survey to capture travel mode choices among users in Wuhan, China across realistic travel scenarios. Attributes for each mode in the survey are derived from actual operational data, ensuring realistic representation. The collected survey data undergo explanatory factor analysis to uncover underlying latent factors that shape traveler preferences. Subsequently, we employ SEM, which comprises both measurement and structural models, to systematically explore the latent psychological constructs such as perceived reliability, social influence, trust, performance expectations, and policy support, and their interrelationships influencing behavioral intentions to adopt AR services. This comprehensive methodological approach enables an accurate identification of key psychological drivers behind mode choice decisions, supporting robust policy recommendations and effective marketing strategies for enhancing AT adoption.

There are two key components in the SEM, namely the measurement model and the structural model. The measurement model specifies how observed variables (indicators) represent latent psychological constructs:

\begin{linenomath}
  \begin{equation}
 X = \Lambda_X \xi + \delta
  \end{equation}
\end{linenomath}

\begin{linenomath}
  \begin{equation}
Y = \Lambda_Y \eta + \epsilon
  \end{equation}
\end{linenomath}

where $X$ and $Y$ represent vectors of observed indicators for the exogenous and endogenous latent constructs, respectively, $\xi$ represents the vector of exogenous latent constructs (e.g., perceived reliability, social influence), $\eta$ represents the vector of endogenous latent constructs (e.g., behavioral intention), $\Lambda_X$ and $\Lambda_Y$ are factor loading matrices linking observed indicators to their respective latent constructs, and $\delta$ and $\epsilon$ are vectors of measurement errors associated with observed indicators.

Meanwhile, the structural model specifies the relationships among latent constructs as follows:

\begin{linenomath}
  \begin{equation}
\eta = B\eta + \Gamma \xi + \zeta
  \end{equation}
\end{linenomath}

where $B$ represents the coefficient matrix depicting relationships among endogenous latent constructs, $\Gamma$ is the coefficient matrix depicting the direct influences of exogenous latent constructs on endogenous latent constructs, and $\zeta$ is the structural error term representing unexplained variance in the endogenous latent constructs.

\section{Survey Questionnaire Design and Data Collection}
\label{sec: survey}

This section elaborates our survey questionnaire design, implementation, and data collection process.

\subsection{Survey Questionnaire Design}
\begin{figure}[!ht]
  \centering
  \includegraphics[width=0.6\textwidth]{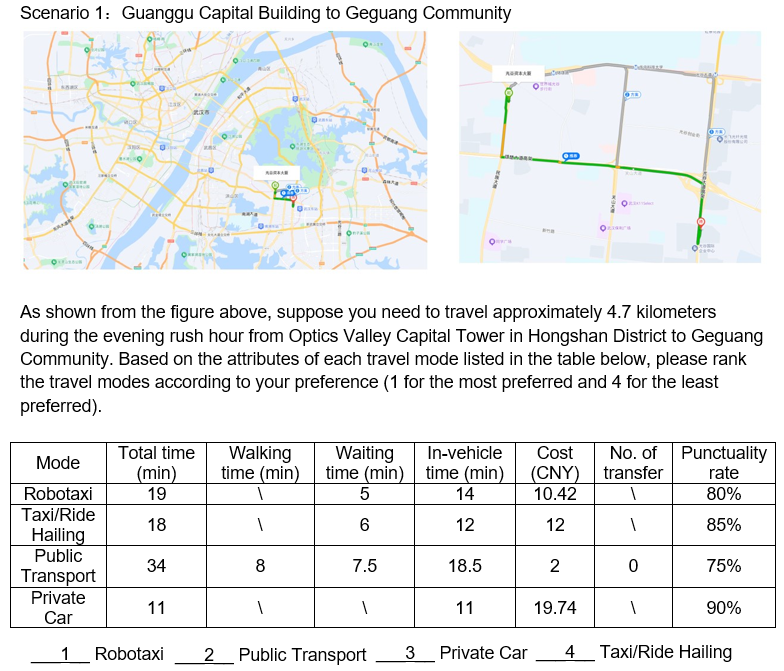}
  \caption{Sample Travel Scenario involving AR in Wuhan, China}\label{fig: exampleScenario}
\end{figure}

A SP survey was designed using Qualtrics to investigate travelers' perceptions and preferences regarding different travel mode choices, including the newly introduced yet fully operational AR service. The questionnaire consists of three main sections. The first section collects respondents' demographic information and their prior experience with AR. In the second section, respondents rank their preferred travel alternatives across ten realistic scenarios, involving trip distances ranging from 4 km to 18 km. Attributes for these scenarios—including waiting time, in-vehicle travel time, and costs—were derived directly from real-world operational data. Ride-hailing and AR service attributes were obtained from Didi and AR apps during evening peak hours, whereas public transport and private car attributes were sourced from Baidu Map API recommendations. The final section consists of open-ended questions to capture additional qualitative insights into respondents' attitudes toward AR services. An example of a travel scenario is presented in Figure~\ref{fig: exampleScenario}.

To ensure questionnaire quality and clarity, a pilot survey was conducted online with professionals, who provided feedback on aspects including question clarity, scenario realism, completion duration, and potential misunderstandings. Based on the feedback, the questionnaire was further refined to enhance its clarity, realism, and objectivity. The complete questionnaire is available in the Appendix.

\subsection{Survey Execution and Data Collection}
To obtain a comprehensive understanding of traveler preferences, a hybrid survey combining both on-site and online stated preference (SP) methods was conducted in Wuhan, China. The online survey allowed for extensive reach and a greater number of responses, whereas the on-site survey enabled targeted sampling of user groups typically underrepresented in online surveys. This combined approach not only increased the total sample size but also improved demographic balance and representativeness, thereby enhancing the robustness of the subsequent explanatory factor analysis and structural equation analysis.

\begin{table}[!ht]
\centering
\caption{Demographic information of survey respondents \label{table: demographics}}

\small
\renewcommand\arraystretch{1.35}{
\begin{tabular}{l l l l l }
\toprule
{\bf Attributes}   & {\bf Category} & {\bf Percentage} & {\bf Category} & {\bf Percentage} \\ \midrule
\multicolumn{1}{c}{Gender}  & Male & 53.6\% & Female & 46.4\% \\ \cline{2-5} 
\multicolumn{1}{c}{Age}  & Below 20  & 4.5\%   & 20--29    & 38.7\%   \\
 & 30--39    & 44.0\%   & 40--49   & 8.6\%   \\
 & Above 50 & 4.2\%    &    &  \\\cline{2-5}
\multicolumn{1}{c}{Education}   & Secondary School   & 3.3\% & Junior College & 8.6\%      \\
 & Diploma    & 24.1\%  & Bachelor    & 51.2\%   \\
  & Master or PhD     & 11.9\%   & Others   & 0.9\%  \\ \cline{2-5}
\multicolumn{1}{c}{No. of Family Members}  & $\leq 3$ & 48.2\% & 4 or 5   & 44.6\% \\
 & $\geq 6$ & 7.2\%  &   &  \\ \cline{2-5}
\multicolumn{1}{c}{Household Monthly Income (CNY)}   & \textless{}5k  & 4.8\%   & 5k to 8k  & 8.9\%  \\
  & 8k to 12k    & 18.4\%   & 12k to 18k   & 31.8\%   \\
  & 18k to 25k    & 24.7\%   & 25k to 40k   & 8.7\%  \\
  & \textgreater{}40k   & 2.7\%      &      &    \\ \cline{2-5}
\multicolumn{1}{c}{City of Origin}   & Wuhan  & 92.6\%  & Others   & 7.4\%   \\ \cline{2-5}
\multicolumn{1}{c}{Car Ownership}   & Yes & 69.9\%   & No   & 30.1\% \\ \cline{2-5}
\multicolumn{1}{c}{Driving Experience}   & Yes & 77.0\%   & No   & 23.0\% \\ \cline{2-5}

\multicolumn{1}{c}{Experience with ARs before}     & Yes   & 60.1\%     & No  & 39.9\% \\    

\bottomrule
\end{tabular}
}

\end {table}

The on-site SP survey was conducted between October 17th and October 21st, 2024, across more than 40 designated AR pick-up and drop-off points in Wuhan, where 600 questionnaires were distributed, resulting in 221 responses. The online survey, conducted from March 8th to March 17th, 2025, garnered 211 responses. In total, 811 questionnaires were disseminated, yielding 432 returned responses. Following a careful screening process to remove inattentive responses completed in less than seven minutes, 336 complete and valid questionnaires remained for further analysis. Participants required approximately 10–12 minutes on average to complete the survey and were each compensated 15 CNY to encourage thoughtful participation and maintain data quality. It is noteworthy that traditional ride-hailing drivers exhibited significant reluctance to participate, perceiving AR services as direct competitors, resulting in a response rate of only about 20\%, compared to approximately 50\% for other respondent groups. The demographic details of survey respondents are summarized in Table~\ref{table: demographics}.

As shown in Table~\ref{table: demographics}, the demographic characteristics of respondents closely match those of the general population in Wuhan, based on official statistics (\url{https://tjj.wuhan.gov.cn/ztzl\_49/pczl/202109/t20210916\_1779157.shtml}, in Chinese). Although the sample slightly favors male respondents, the gender distribution remains relatively balanced. Given that AR technology represents an emerging high-tech innovation, it is unsurprising that more than 80\% of respondents were young adults aged between 20 and 39 years, with approximately 63.1\% holding a bachelor's, master's, or doctoral degree—demographics more likely to embrace new mobility technologies. The average monthly individual income reported by respondents is approximately CNY 7,643, slightly lower than Wuhan's official average salary of CNY 9,178 in 2023 (\url{https://tjj.wuhan.gov.cn/}, \url{https://teamedupchina.com/average-salary-in-wuhan-china/}), likely due to the inclusion of lower-income or unemployed participants, such as university students. Although the survey was primarily conducted in Wuhan, a small portion of respondents were from other cities, potentially tourists, migrant workers, or AR developers. Additionally, over 60\% of respondents reported prior experience with AR services, reflecting the extensive deployment of ARs in Wuhan since early 2024, thus providing valuable insights grounded in real-world usage. Consequently, the survey responses effectively represent local perceptions and attitudes toward adopting AR services as a standard mode of urban mobility.

The responses from the survey respondents for each open question are summarized in the table below. It is worth noting that for each respondent, we record number of times they choose AR as the travel mode in the 10 typical realistic travel scenarios in the city of Wuhan. Together, they will be used in the explanatory factor analysis to set up the measurement model and derive the path diagram of the structural model in the structural equation analysis.

\begin{table}[ht!]
\centering
\small
\caption{Survey Questions and Likert Scale Response Options}
\label{tab:survey_questions}
\begin{tabular}{l l l l l l l}
\hline
\textbf{No.} & \textbf{Survey Questions} & \textbf{Strongly}  & \textbf{Disagree} & \textbf{Neutral} & \textbf{Agree} & \textbf{Strongly} \\ 
 & &  \textbf{Disagree} & & & & \textbf{Agree}  \\
\hline
Q1 & I think ATs are safe to use. &0.66\%&4.26\%&29.51\%&50.82\%&14.75\%

 \\ 
\hline
Q2 & ATs are more convenient than \\& traditional taxis or ride-hailing services. &1.64\%&9.51\%&32.46\%&33.11\%&23.28\%
\\ 
\hline
Q3 & I'm willing to commute daily using ATs. &1.31\%&6.23\%&26.56\%&47.54\%&18.36\%
\\ 
\hline
Q4 & I enjoy the technological feel of riding in ATs. &0.33\%&6.56\%&21.31\%&43.93\%&27.87\%
\\ 
\hline
Q5 & I'd continue frequently using ATs even if fares\\&  matched or slightly exceeded traditional taxis. &8.52\%&25.9\%&29.18\%&24.59\%&11.8\%
\\ 
\hline
Q6 & I trust that AT technology is reliable and stable. &0.33\%&5.57\%&27.87\%&48.85\%&17.38\%
\\ 
\hline
Q7 & The operation of AT technology is transparent \\& and understandable. &0.98\%&8.85\%&24.92\%&45.25\%&20\%
\\ 
\hline
Q8 & I trust AT technology securely handles my data. &0.66\%&7.54\%&26.89\%&39.34\%&25.57\%
\\ 
\hline
Q9 & Friends and family's use of ATs positively \\& influences my choice. &1.31\%&14.75\%&25.25\%&43.28\%&15.41\%
\\ 
\hline
Q10 & Media reports enhance my trust in AT technology. &1.64\%&10.49\%&22.62\%&44.59\%&20.66\%
\\ 
\hline
Q11 & Seeing widespread AT use increases my trust.&0.33\%&5.57\%&17.38\%&48.2\%&28.52\%
\\ 
\hline
Q12 & I'm willing to try ATs. &0.33\%&3.28\%&16.39\%&53.77\%&26.23\%
\\ 
\hline
Q13 & I plan to regularly use ATs in daily life. &1.64\%&10.49\%&28.85\%&38.69\%&20.33\%
\\ 
\hline
Q14 & I'd recommend ATs to others. &0.66\%&7.87\%&26.23\%&44.26\%&20.98\%
\\ 
\hline
Q15 & I support government funding for promoting ATs. &1.64\%&2.62\%&17.7\%&50.49\%&27.54\%
\\ 
\hline
Q16 & Policies should be designed to encourage \\& widespread AT adoption. &1.64\%&3.61\%&18.36\%&51.8\%&24.59\%
\\ 
\hline
Q17 & Regulations should ensure AT technology's safe \\& implementation. &0\%&3.61\%&11.48\%&47.54\%&37.38\%
\\ 
\hline
Q18 & Current legislation and oversight of ATs are \\& insufficient and need strengthening. &3.28\%&4.92\%&15.08\%&42.62\%&34.1\%
\\ 
\hline
Q19 & AT rides are comfortable. &0.98\%&1.64\%&23.61\%&50.16\%&23.61\%
\\ 
\hline
Q20 & Riding in an AT is reassuring. &0.66\%&9.18\%&25.25\%&40.98\%&23.93\%
\\ 
\hline
Q21 & AT fares are affordable. &0.66\%&6.56\%&34.75\%&41.31\%&16.72\%
\\ 
\hline
Q22 & ATs offer good privacy. &0.98\%&8.2\%&23.28\%&41.31\%&26.23\%
\\ 
\hline
Q23 & ATs can reliably reach destinations on time. &1.64\%&6.56\%&25.25\%&44.92\%&21.64\%
\\ 
\hline
Q24 & Waiting times for ATs are reasonable. &1.97\%&9.18\%&24.92\%&44.92\%&19.02\%
\\ 
\hline
Q25 & ATs enable quick travel to destinations. &1.31\%&12.79\%&24.26\%&41.64\%&20\%
\\ 
\hline
\end{tabular}
\end{table}

Based on the Likert-scale response data summarized in Table~\ref{tab:survey_questions}, several noteworthy interpretations can be made regarding travelers’ perceptions and attitudes toward the AT service AR in Wuhan. Overall, respondents exhibited a generally favorable disposition toward AT services. For instance, approximately 66\% of participants agreed or strongly agreed that ATs are safe to use, with similarly high agreement observed for items related to technology reliability and stability. Such findings suggest that ATs have achieved a significant level of trust and acceptance among the surveyed population, likely influenced by their real-world operational exposure to the AR service.

A substantial proportion of respondents reported positive perceptions of ATs in terms of ride performance and comfort. Specifically, over 70\% of respondents agreed or strongly agreed that AT rides are comfortable, and more than 65\% felt reassured when riding in them. Furthermore, more than 58\% found the fare levels to be affordable, and a majority of respondents believed that ATs arrived punctually and required only reasonable waiting times. These results indicate that operational features of AT services, such as timeliness, cost, and comfort, are well aligned with user expectations, which enhances user confidence and satisfaction.

Trust in the underlying AV technology was also notably strong among respondents. The majority agreed that the system was reliable, transparent in its operation, and capable of safeguarding user data. Approximately 66\% expressed trust in the reliability and stability of ATs, while similar levels of agreement were observed for statements related to data privacy and system comprehensibility. These items collectively point toward a well-established sense of technological trust, which is likely to play a central role in shaping behavioral intentions in the structural equation analysis.

Social influence and media exposure also emerged as important contextual factors. Respondents reported that their decisions were influenced by the usage behavior of friends and family, and more than 76\% indicated that seeing more people using ATs increased their own trust in the technology. In addition, a majority of respondents stated that favorable media coverage improved their perception of ATs. These results suggest that social norms and external communication channels significantly contribute to public perception, and these factors should be modeled as latent constructs capturing social influence dynamics.

Respondents also demonstrated strong behavioral intention to adopt ATs. Over 80\% were willing to try AT services, 59\% planned to use them regularly, and approximately 65\% would recommend them to others. These responses provide robust evidence for the inclusion of an intention-to-use construct in the SEM, supported by high levels of engagement and openness to adoption.

Policy-related items revealed substantial public support for governmental involvement in the promotion and regulation of ATs. Approximately 78\% supported government subsidies and resource allocation, and an even higher proportion agreed that regulations should be implemented to ensure safe deployment. Notably, around 77\% believed that current regulations and oversight were insufficient and required improvement. This reflects a clear public expectation for a stronger regulatory framework and active policy backing, which can serve as a critical enabler for the expansion of AV-based services.

Despite these generally positive findings, some reservations about pricing were evident. When asked whether they would continue using ATs if prices rose to match or slightly exceed those of traditional services, only 36.4\% responded affirmatively, while approximately 34\% expressed disagreement. This indicates that price competitiveness remains a potential barrier to broader adoption, and suggests the importance of fare design or targeted subsidies to ensure continued usage.

\begin{figure}[!ht]
  \centering
  \includegraphics[width=0.8\textwidth]{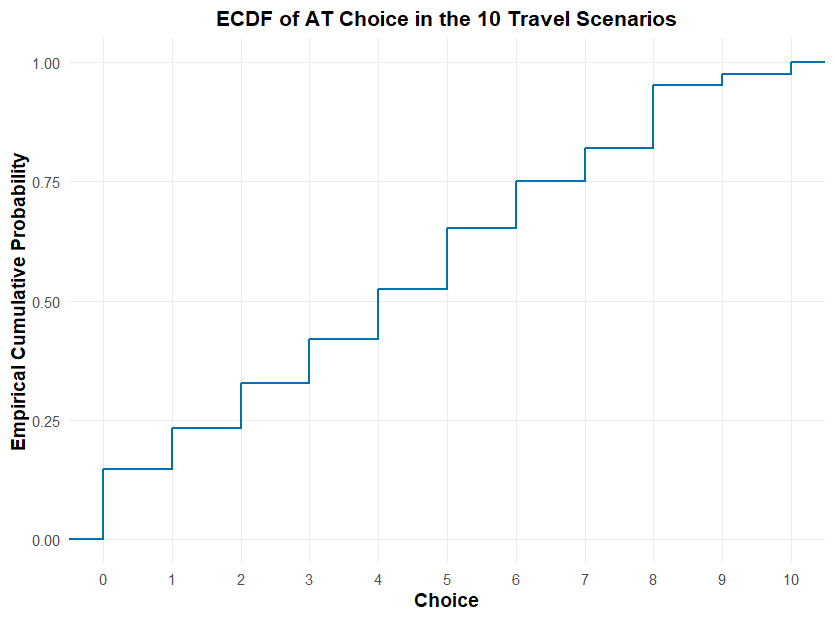}
  \caption{ECDF of AT choice in the 10 travel scenarios}\label{fig: ECDF}
\end{figure}

Figure~\ref{fig: ECDF} presents the empirical cumulative distribution function (ECDF) of AT selection across the ten travel scenarios in the SP survey. Each respondent could select ATs as their top travel mode up to ten times, depending on their preferences for each scenario. The ECDF curve reveals a steady increase in cumulative probability with the number of AT selections, indicating a diverse range of user attitudes toward AT adoption. Approximately 50\% of respondents chose ATs as their preferred mode in at least five out of ten scenarios, reflecting a moderate-to-high level of acceptance. Moreover, a noticeable fraction of respondents selected ATs for most or all scenarios, suggesting that certain user segments are already highly receptive to integrating AT services into their regular mobility behavior. This cumulative pattern highlights meaningful adoption tendencies and provides an observed independent behavioral variable for use in the SEM framework to uncover the latent psychological constructs underlying travel mode choices.

Taken together, these descriptive results provide empirical justification for the SEM framework employed in this study. The variation across responses suggests the presence of distinct latent psychological constructs including perceived performance, trust, social influence, behavioral intention, and policy support that can shape the adoption of ATs. These constructs, along with the observed behavioral variable representing the number of times respondents chose AT in the stated preference scenarios, will form the basis of the measurement and structural models, enabling a comprehensive analysis of the psychological and behavioral mechanisms underlying the adoption of autonomous mobility services.

\section{Results}
\label{sec: results}
This section presents an in-depth analysis of survey responses to identify the latent factors. A explanatory factor analysis is first performed, followed by the building of the measurement model of the SEM. Subsequently, the latent factors from the measurement model are used to be associated with the AT adoption behavior, which is the frequency of choosing AR in the 10 representative travel scenarios in Wuhan. The dataset for the following analyses consists of 3,36 observations, derived from 336 eligible responses.

\subsection{Explanatory Factor Analysis}
Explanatory Factor Analysis (EFA) was conducted to uncover the underlying latent constructs that account for the variability among the observed survey responses. Utilizing the \texttt{psych} package in \texttt{R}, EFA was applied to the attitudinal variables and socio-demographic variables to determine the appropriate number of latent factors for subsequent structural modeling. Table~\ref{tab:efa_loadings} presents the factor loadings and variance explained from the EFA , where the results indicate a six-factor solution, accounting for approximately 36.8\% of the total variance. For clarity, only those loadings that are > 0.3 are displayed in the table. Among them, the first latent factor (ML1) explains the largest share of the variance (12.1\%), followed by ML5 (7.1\%), ML6 (5.7\%), ML4 (3.7\%), ML3 (4.6\%), and ML2 (3.6\%).

\begin{table}[ht!]
\centering
\caption{EFA Loadings}
\label{tab:efa_loadings}
\small
\begin{tabular}{lllllll}
\hline
\textbf{Item} & \textbf{ML1} & \textbf{ML5} & \textbf{ML6} & \textbf{ML4} & \textbf{ML3} & \textbf{ML2} \\
\hline
Q1  & 0.544 &       &       &       &       &       \\
Q2  &       & 0.386 &       &       &       &       \\
Q3  &       &       & 0.313 &       &       &       \\
Q4  & 0.688 &       &       &       &       &       \\
Q5  &       &       &       & 0.666 &       &       \\
Q6  & 0.487 &       &       &       &       &       \\
Q7  & 0.490 &       &       &       &       &       \\
Q8  & 0.475 &       &       &       &       &       \\
Q9  &       &       & 0.560 &       &       &       \\
Q10 &       &       &       &       &       &       \\
Q11 & 0.493 &       &       &       &       &       \\
Q12 & 0.399 &       & 0.378 &       &       &       \\
Q13 &       &       & 0.458 &       &       &       \\
Q14 &       &       & 0.660 &       &       &       \\
Q15 & 0.834 &       &       &       &       &       \\
Q16 & 0.646 &       &       &       &       &       \\
Q17 & 0.463 &       &       &       &       &       \\
Q18 &       &       &       & -0.457 &      &       \\
Q19 & 0.540 &       &       &       &       &       \\
Q20 & 0.338 &       & 0.354 &       &       &       \\
Q21 &       & 0.410 &       &       &       &       \\
Q22 &       & 0.345 &       &       &       &       \\
Q23 &       & 0.703 &       &       &       &       \\
Q24 &       & 0.562 &       &       &       &       \\
Q25 &       & 0.769 &       &       &       &       \\
Income             &       &       &       &       &       & 0.393 \\
Gender             &       &       &       &       &       &       \\
Car Ownership       &       &       &       &       & 0.726 &       \\
Driving Experience  &       &       &       &       & 0.713 &       \\
Education          &       &       &       &       &       & 0.974 \\
Age                &       &       &       &       & 0.402 &       \\
\hline
\textbf{SS Loadings}     & 4.001 & 2.349 & 1.885 & 1.219 & 1.512 & 1.176 \\
\textbf{Proportion Var}  & 0.121 & 0.071 & 0.057 & 0.037 & 0.046 & 0.036 \\
\textbf{Cumulative Var}  & 0.121 & 0.192 & 0.250 & 0.286 & 0.332 & 0.368 \\
\hline
\end{tabular}
\end{table}

ML1 appears to capture general trust and confidence in ATs, with high loadings from items such as “I think ATs are safe to use” (Q1), “I enjoy the technological feel of riding in ATs” (Q4), and strong support for policy-related measures (Q15–Q17). Items such as “I trust AT technology securely handles my data” (Q8) and “I trust that AT technology is reliable and stable” (Q6) further support the interpretation of this factor as capturing perceptions of safety, technological assurance, and policy support.

ML2 is characterized by a single dominant loading on education (0.974), identifying it as a demographic factor reflecting the respondent’s educational attainment, which may indirectly influence perceptions or behavioral intentions toward AT usage.

ML3 consists mostly of socio-demographic variables such as income, car ownership, driving experience, and age, suggesting that this factor captures mobility-related lifestyle characteristics. Notably, car ownership and driving experience load very strongly here (0.726 and 0.713, respectively), while income and age load moderately, indicating this factor’s role in reflecting users' existing travel habits and capacity to adapt to new technologies.

ML4 seems to represent attitudinal resistance or cost sensitivity, as it includes a strong positive loading on Q5 (continued usage despite fare increase) and a negative loading on Q18 (perceived inadequacy of current legislation). This dimension may reflect the willingness to continue using ATs under varying policy and pricing conditions.

ML5 is primarily associated with perceived operational performance, with strong loadings from Q23–Q25, reflecting perceptions of timeliness, waiting time, and speed of AT services. Moderate loadings on Q2 (convenience) and Q21–Q22 (affordability and privacy) suggest that this factor represents how well the service performs in practice relative to user expectations.

ML6 is interpretable as behavioral intention and social influence, indicated by relatively strong loadings on Q9 (peer usage), Q13–Q14 (regular use and recommendations), and Q12 (willingness to try ATs). These items reflect users' readiness to adopt ATs and the influence of social surroundings on their attitudes.

Overall, the factor structure aligns well with the theoretical constructs underpinning the measurement model in the SEM framework. Each latent dimension is well represented by a coherent set of observed variables, supporting construct validity and the subsequent specification of the structural model for analyzing adoption behavior.

\subsection{Structural Equation Modeling}

\begin{figure}[!ht]
  \centering
  \includegraphics[width=1\textwidth]{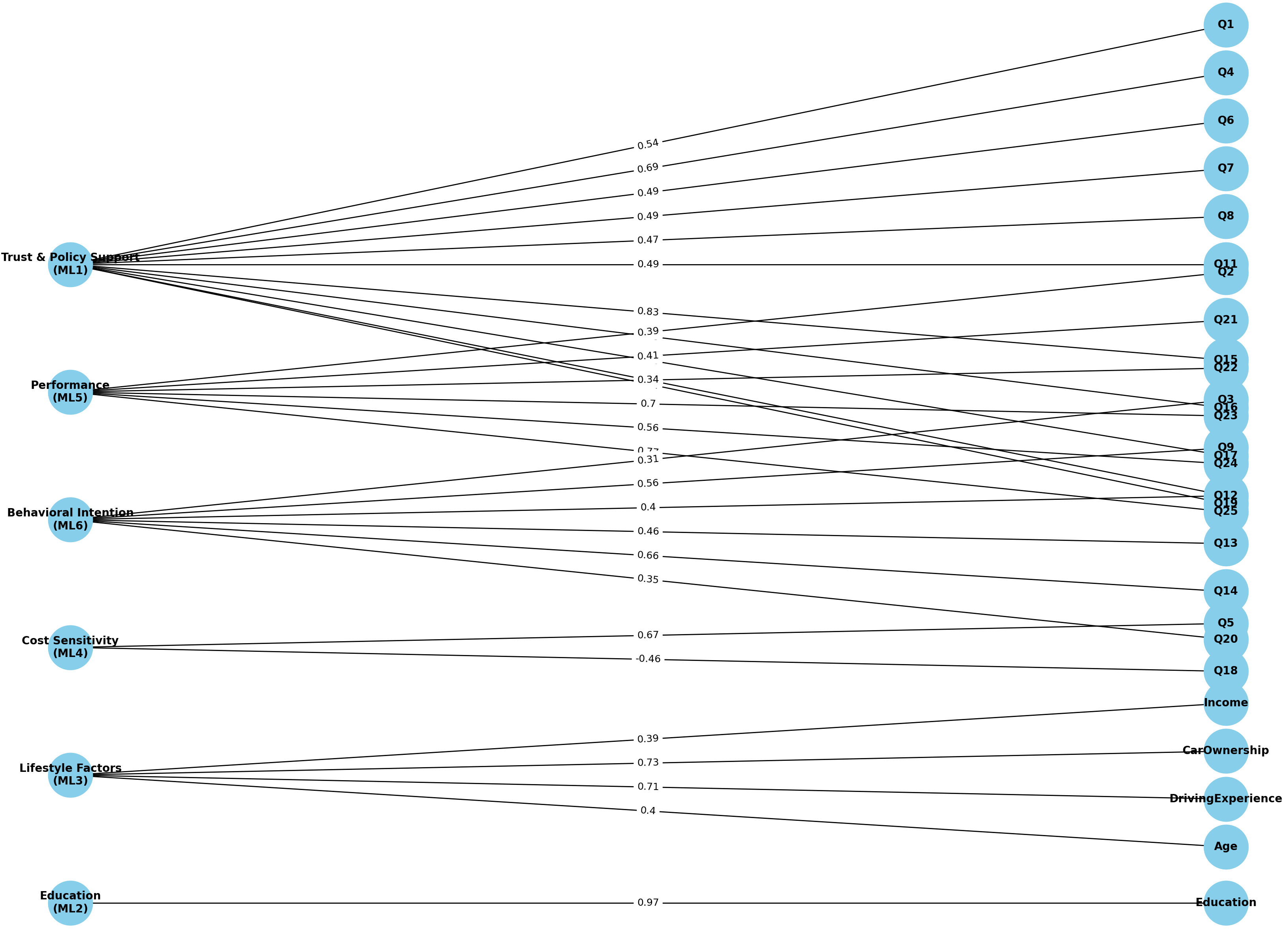}
  \caption{The measurement model with loading factors}\label{fig: measurement}
\end{figure}

Following the results of the EFA, the measurement model is developed as the first component of the SEM procedure. Figure~\ref{fig: measurement} presents the measurement model, which includes both the observed variables (rectangles) and the latent constructs (ellipses). The arrows from the latent constructs to the observed indicators represent factor loadings. The constructs are allowed to covary, as indicated by the bidirectional arrows connecting the ellipses. As seen in the figure, the majority of indicators load strongly (i.e., standardized loadings $> 0.7$) onto their respective latent constructs, indicating adequate convergent validity.

\begin{figure}[!ht]
  \centering
  \includegraphics[width=1\textwidth]{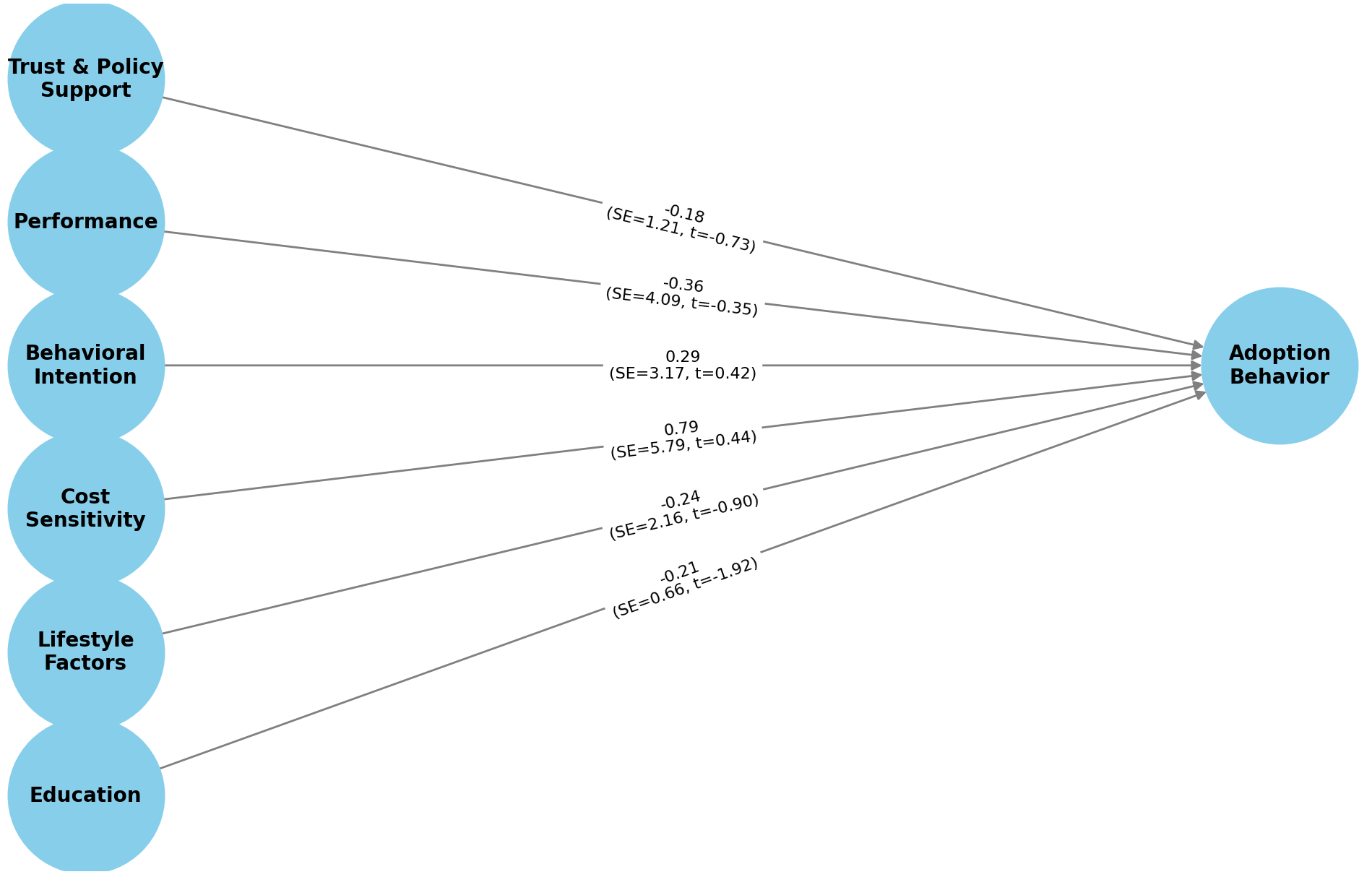}
  \caption{The structural model for AT adoption}\label{fig: structural}
\end{figure}

Once the measurement model was validated, the structural model was estimated to investigate the effect of latent psychological constructs on individuals’ AT adoption behavior by looking at their number of AT choices in the 10 representative travel scenarios in the survey questionnaire. Figure~\ref{fig: structural} illustrates the structural model, where the dependent variable is Adoption Behavior and the six latent constructs are modeled as predictors. The path coefficients are presented as standardized estimates, with their respective standard errors (SE) and $t$-values reported in parentheses below each coefficient path.

As shown in Figure~\ref{fig: structural}, the construct \textit{Cost Sensitivity} exhibited the strongest positive association with adoption behavior ($\beta = 0.79$, SE = 5.79, $t = 0.44$), suggesting that perceived economic benefits significantly influence willingness to adopt autonomous taxis. Similarly, \textit{Behavioral Intention} showed a moderate positive effect ($\beta = 0.29$, SE = 3.17, $t = 0.42$). In contrast, constructs such as \textit{Trust \& Policy Support}, \textit{Performance}, \textit{Lifestyle Factors}, and \textit{Education} yielded negative path coefficients.

Table~\ref{tab:model_fit} summarizes the key model fit indices for the structural model. The comparative fit index (CFI) and Tucker-Lewis index (TLI) were 0.956 and 0.950, respectively, both meeting the recommended threshold of $\geq 0.95$ for a good model fit. The root mean square error of approximation (RMSEA) was 0.043 with a 90\% confidence interval of [0.036, 0.050], falling well within the acceptable range. Furthermore, the standardized root mean square residual (SRMR) was 0.046, which is below the 0.08 benchmark. These indices collectively indicate that the overall model exhibits a good fit to the data.

In summary, the SEM analysis suggests that all six latent factors contribute conceptually to understanding adoption behavior, and cost sensitivity and behavioral intention impose more significant influences. These findings can inform both policymakers and service providers on which psychological factors to target in promoting AV adoption.

\begin{table}[ht!]
\centering
\caption{Fitness metrics of the SEM}
\label{tab:model_fit}
\begin{tabular}{lll}
\hline
\textbf{Fit Index} & \textbf{Value} & \textbf{Recommended Threshold} \\
\hline
Comparative Fit Index (CFI) & 0.956 & $\geq 0.95$ (good) \\
Tucker-Lewis Index (TLI) & 0.950 & $\geq 0.95$ (good) \\
Root Mean Square Error of Approximation (RMSEA) & 0.043 & $\leq 0.06$ (good) \\
90\% CI RMSEA (Lower) & 0.036 & CI lower bound close to 0 \\
90\% CI RMSEA (Upper) & 0.050 & CI upper bound $< 0.08$ \\
Standardized Root Mean Square Residual (SRMR) & 0.046 & $< 0.08$ (good) \\
\hline
\end{tabular}
\end{table}

\section{Conclusions}

\section{Conclusion}
This study investigates the adoption behavior of AT services by leveraging real-world exposure to Baidu's AR service in Wuhan, China. Unlike prior research relying on hypothetical scenarios, we design a SP survey grounded in actual service attributes and behavioral outcomes. Through EFA and SEM, we identify and quantify six key latent psychological constructs, namely Trust \& Policy Support, Cost Sensitivity, Performance, Behavioral Intention, Lifestyle, and Education. We also explore their influence on adoption behavior, defined as the frequency of AT selections across ten realistic travel scenarios.

Our findings highlight that Cost Sensitivity and Behavioral Intention exert the strongest positive influences on AT adoption. Specifically, travelers who perceive ATs as economically advantageous or express a willingness to integrate ATs into daily routines are more likely to adopt these services. Other factors, such as trust in the technology, policy support, and socio-demographic variables, show indirect or weaker effects, underscoring the importance of economic incentives and personal readiness in shaping adoption decisions.

The model exhibits excellent fit, validating the conceptual framework and supporting the robustness of our SEM approach. These insights offer valuable guidance for policymakers and mobility providers: aggressive fare subsidies, improved service visibility, and targeted communication campaigns that emphasize reliability and user experiences may significantly enhance public acceptance. Moreover, regulatory clarity and user trust-building initiatives should accompany technological deployment to strengthen long-term adoption.

Future research may explore longitudinal user behavior, cross-city comparisons, or hybrid modeling that integrates revealed and stated preferences. As AV-based mobility services transition from pilot programs to full-scale deployment, understanding the psychological and behavioral drivers of adoption becomes increasingly critical for shaping sustainable and inclusive urban transportation futures.

\section{Acknowledgements}
The research is supported by the research grant from the National Natural Science Foundation Council of China (Grant Nos. 72471070). Any opinions, findings, and conclusions or recommendations expressed in this paper are those of the authors and do not necessarily reflect the views of the sponsors.

\section{Appendix}
\label{sec: appendix}
The complete SP survey questionnaire can be accessed at \url{https://drive.google.com/drive/folders/18Lk22sODYZlw7GjZgaC1bb\_4ky2DveAo?usp=sharing}.

\newpage

\bibliographystyle{trb}
\bibliography{trb_template}
\end{document}